\documentclass[journal]{IEEEtran}
\pdfoutput=1 
\usepackage{times}
\usepackage{epsfig}
\usepackage{graphicx}
\usepackage{amsmath}
\usepackage{amssymb}
\usepackage{subfigure}
\usepackage{amsfonts}
\usepackage{multirow}
\usepackage[pagebackref=true,breaklinks=true,letterpaper=true,colorlinks,bookmarks=false]{hyperref}

\usepackage{tikz}
\usepackage{url}

\newcommand{\pushright}[1]{\ifmeasuring@#1\else\omit\hfill$\displaystyle#1$\fi\ignorespaces}
\newcommand{\pushleft}[1]{\ifmeasuring@#1\else\omit$\displaystyle#1$\hfill\fi\ignorespaces}

\title{Multi Channel-Kernel Canonical Correlation Analysis for Cross-View Person Re-Identification}

\author{Giuseppe Lisanti, Svebor Karaman, Iacopo Masi\thanks{G. Lisanti is with the Media Integration and Communication Center (MICC)
of the University of Florence, Florence, Italy. E-mail: \{firstname.lastname\}@unifi.it.}\thanks{S. Karaman is with the Digital Video and Multimedia Lab, Department of Electrical Engineering of Columbia University, New York, NY, USA. E-mail: svebor.karaman@columbia.edu}\thanks{I. Masi is with the University of Southern California, Los Angeles, CA, USA. E-mail: iacopo.masi@usc.edu}\thanks{Manuscript received \today.}}

\begin{document}
\maketitle
\begin{tikzpicture}[remember picture, overlay]\node[yshift=-0.5cm] at (current page.north) {This is a pre-print version, the final version of the manuscript with more experiments can be found at:};\node[yshift=-1cm] at (current page.north) {\url{https://doi.org/10.1145/3038916}};\end{tikzpicture}

\maketitle

\begin{abstract}
In this paper we introduce a method to overcome one of the main challenges of person re-identification in multi-camera networks, namely cross-view appearance changes. The proposed solution addresses the extreme variability of person appearance in different camera views by exploiting multiple feature representations. For each feature, Kernel Canonical Correlation Analysis (KCCA) with different kernels is exploited to learn several projection spaces in which the appearance correlation between samples of the same person observed from different cameras is maximized. An iterative logistic regression is finally used to select and weigh the contributions of each feature projections and perform the matching between the two views. Experimental evaluation shows that the proposed solution obtains comparable performance on VIPeR and PRID 450s datasets and improves on PRID and CUHK01 datasets with respect to the state of the art.
\end{abstract}

\begin{IEEEkeywords}
person re-identification, KCCA, late fusion
\end{IEEEkeywords}

\IEEEpeerreviewmaketitle

\section{Introduction}
Video surveillance systems are now ubiquitous in public areas such as airports, train
stations or even city wide. These systems are typically implemented in the form of camera networks and cover very large areas, with limited or no overlap between different camera views. They should be able to
track a person throughout the network, matching detections of the same person in different camera views, irrespectively of view and illumination changes, as well as pose and scale variations of the person. 
Matching person detections across a camera network is
typically referred as \textit{re-identification}.

In this paper, we propose a solution for person re-identification that grounds on the idea of addressing the extreme variability of person appearance in different camera views through a multiplicity of representations projected onto multiple spaces that emphasize appearance correlation and finally learning the most appropriate combinations for the observed pair. 
In particular, several texture and color features are extracted from a coarse segmentation of the person image to account for viewpoints and illumination changes.   For each feature, we learn several projection spaces where features from two cameras correlate,  using Kernel Canonical Correlation Analysis (KCCA) with different kernels. Finally, matching between two views is obtained by exploiting logistic regression to weigh the contributions of feature projections.

\subsection{Related works}
\label{sec:relatedworks}
Re-identification has been an active subject of research for several years. We review the most important works in the following.

\subsubsection{Methods based on hand-crafted descriptors}
\label{sec:appearance}
These methods concentrate on the definition of descriptors that are able to capture as much as possible the variability of person appearance in different views. 
Among the best performing proposals in this class,  the Symmetry-Driven Accumulation of Local Features (SDALF) descriptor~\cite{bazzani:reidcvpr10} takes into account image segments of physical parts of the human body such as the head, torso, and legs, obtained from the computation of axis symmetry and asymmetry and background modeling. 
For each segment, color information is represented by weighted HSV color histograms and maximally stable color regions,
and texture information is encoded as recurrent highly-structured patches. 
In~\cite{ChengBMVC11} the same authors proposed to fit a 
Custom Pictorial Structure
(CPS) model on a person detection estimating the head, chest, thighs
and legs positions. Each part is then described by a HSV color histogram and Maximally Stable Color Regions (MSCR). 

Such part-based models, while well performing for ideal camera takes, have poor performance in real scenarios, due to the fact that image quality is often low so that it is hard to precisely detect body parts.

\subsubsection{Methods based on Deep Convolutional Neural Networks}
\label{sec:deep}

As opposed to the design of hand-crafted features, some authors
have exploited Deep Convolutional Neural
Networks (CNN) to build a
representation that captures the variability of person appearance across views.  
One of the first re-identification works  in this class was~\cite{YiDeep:ICPR14}. Successively, in~\cite{YiLL14:arxiv}, the same authors improved their solution 
 by employing a CNN 
 in a ``siamese'' configuration to jointly learn the color
feature, texture feature and the distance function in a unified
framework (Improved DML). 
Similar to~\cite{YiLL14:arxiv}, Ahmed \emph{et
 al.}~\cite{Ahmed2015:CVPR} proposed a siamese deep network
architecture that learns jointly feature representation and terminates
with a logistic regression loss to discriminate between pairs of target
in a same/not-same fashion (Siamese CNN). Finally, Li \emph{et al.}~\cite{Li_2014:CVPR}
used a novel filter pairing neural network (FPNN) with six-layers
to jointly handle photometric and geometric transforms.

While deep learning has had a big impact in general image recognition and recently in face
recognition~\cite{Taigman_2014_CVPR},  the use of Deep Network-based representations for re-identification is negatively affected by low resolution images as usually occur in re-identification contexts and requires the availability of a huge number of person image pairs from different cameras to train a discriminative model.

\subsubsection{Methods based on learning classifiers}
\label{sec:learnMatch}

This class of methods is the most populated and grounds on the idea of learning
classifiers and metrics to recognize persons across views. They currently score the state of the art performance of re-identification.
In~\cite{cvpr12:kissme} the authors
proposed a Mahalanobis based distance learning
that exploits equivalence constraints derived from target labels (KISSME).
The authors of~\cite{EIML:avss12} proposed an impostor-based metric learning method  (EIML), based on a modified version of the Large Margin Nearest Neighbor (LMNN)~\cite{weinberger2009distance} algorithm.
The method in~\cite{xiong2014person} combined Regularized Pairwise Constrained Component Analysis, Kernel Local Fisher Discriminant Analysis, Marginal Fisher Analysis and a Ranking Ensemble Voting Scheme  
with linear, $\chi^2$ and RBF-$\chi^2$ kernels
to extensively evaluate person re-identification performances (KLMM).
Similarly to~\cite{xiong2014person}, the approach in~\cite{SLTRL_Wang}
introduced an explicit non-linear transformation for the original feature space and learned
a linear similarity projection matrix (SLTRL) by maximizing the top-heavy ranking loss
instead of a loss defined by the Area Under the Curve.
Remarkable performance has been also obtained  by~\cite{Paisitkriangkrai_2015_CVPR,ECM_WACV15}. 
The former combined an
ensemble of different distance metric learning approaches, minimizing
different objective functions, while the latter proposed 
a novel ensemble model (ECM) that combines different color descriptors through 
metric learning.

The methods proposed in~\cite{eSDC:cvpr13,zhao2013:iccv13,Zhao2014:CVPR} rely
mainly on dense correspondences and unsupervised learning of features.
In~\cite{eSDC:cvpr13}, a novel method (eSDC) was proposed that applies
adjacency-constrained patch matching to build dense correspondences
between image pairs through a saliency learning method in a
unsupervised fashion.  
The authors of~\cite{zhao2013:iccv13} extended this method by penalizing patches with inconsistent saliency in order to handle misalignment problems (SalMatch).
Finally, instead of relying on hand-crafted features, Zhao \emph{et al.}~\cite{Zhao2014:CVPR}
 proposed to learn mid-level filters
(mFilters). Dense patches are
clustered together in order to create a hierarchical tree, then the
patches inside a node of the tree are used to train a linear SVM that is used to
discriminate patches between two views. Here, the mid filters are represented by the set of SVM weights and biases learned over the nodes. 
Differently from~\cite{Zhao2014:CVPR}, 
the method in~\cite{Shen_2015_ICCV} introduced a structure to encode
cross-view pattern correspondences (CSL) that are used 
jointly with global constraints to exclude spatial misalignments.

The method in~\cite{PLS_prototype} used salient samples from probe and
gallery to build a set of prototypes. These prototypes are used to
weigh the features according to their discriminative power by using
 Partial Least Square (PLS). The final recognition is
performed by fusing different rank results.  In~\cite{Yang2014}
the authors proposed to encode color using color naming. In particular,  color
distributions over color names in different color spaces are 
fused to generate the final feature representation (SCNCD).  At
the end, the method employs the KISSME metric learning framework.
The work~\cite{Shi_2015_CVPR} proposed to
address the person recognition problem relying
on semantic attributes. 
The main underlying idea is that attributes may provide a strong
invariant cue for recognition. Instead of relying on manually labeled
attributes, the model is trained on fashion photography
data. The attributes are learned as latent variables on top of a superpixel representation.
The authors also transferred the learned model to
video-surveillance settings without requiring any surveillance domain
supervision.

Techniques that deal with cross-view matching were proposed in~\cite{Li_2013_CVPR,Liao2015:CVPR}.
These methods are the closest to our approach since they learn feature projections to better perform matching between images of the same person captured from
different cameras. In~\cite{Li_2013_CVPR} the authors 
partitioned the
image spaces of two camera views into different configurations
according to the similarity of cross-view transformations. Then, for each
partition, the visual features of an image pair from different views
are projected to a common feature space and then matched with
softly assigned metrics. The authors of~\cite{Liao2015:CVPR} defined the
LOMO feature which is composed by HSV color histograms over stripes
along with a texture descriptor which improves over the classic
LBP.  Their approach revised the
KISSME metric learning~\cite{cvpr12:kissme} in order to deal with
cross-view matching problem as well.

\subsection{Contributions}

The main contributions of the proposed approach are the following:
\begin{itemize}
\item Differently from~\cite{bazzani:reidcvpr10,ChengBMVC11} our approach for describing a person does not require the detection of image segments that correspond to physical body parts but instead uses a coarse spatial segmentation of the person image into consecutive regions at different heights.
\item For each region we model the variety of the person appearance in different views through a multiplicity of features, similarly to~\cite{lisanti:pami14,karaman2014leveraging}.
\item  For each feature, we learn a set of projection spaces with different kernels, such that images of the same person coming from different cameras are more easily matched. This differs from the approaches of~\cite{Li_2013_CVPR,lisanti:icdsc14,Liao2015:CVPR} where a single projection space was learned. 
\item We use logistic regression to learn weights of projection spaces so that more distinguishing features contribute more to the re-identification and less significant features are dropped out.
\end{itemize}

In the rest of the paper, we expound our person representation, 
in section~\ref{sec:feature}, and discuss in detail the method in section~\ref{sec:MCKCCA} and ~\ref{sec:weightedlatefusion}.  In section~\ref{sec:experiments}, we compare performance for re-identification using KCCA with multiple kernels with respect to using metric learning. We also give an overview of the performance of our method with respect to the state-of-the-art methods of person re-identification.

\begin{figure*}
\centering

\includegraphics[width=0.95\textwidth]{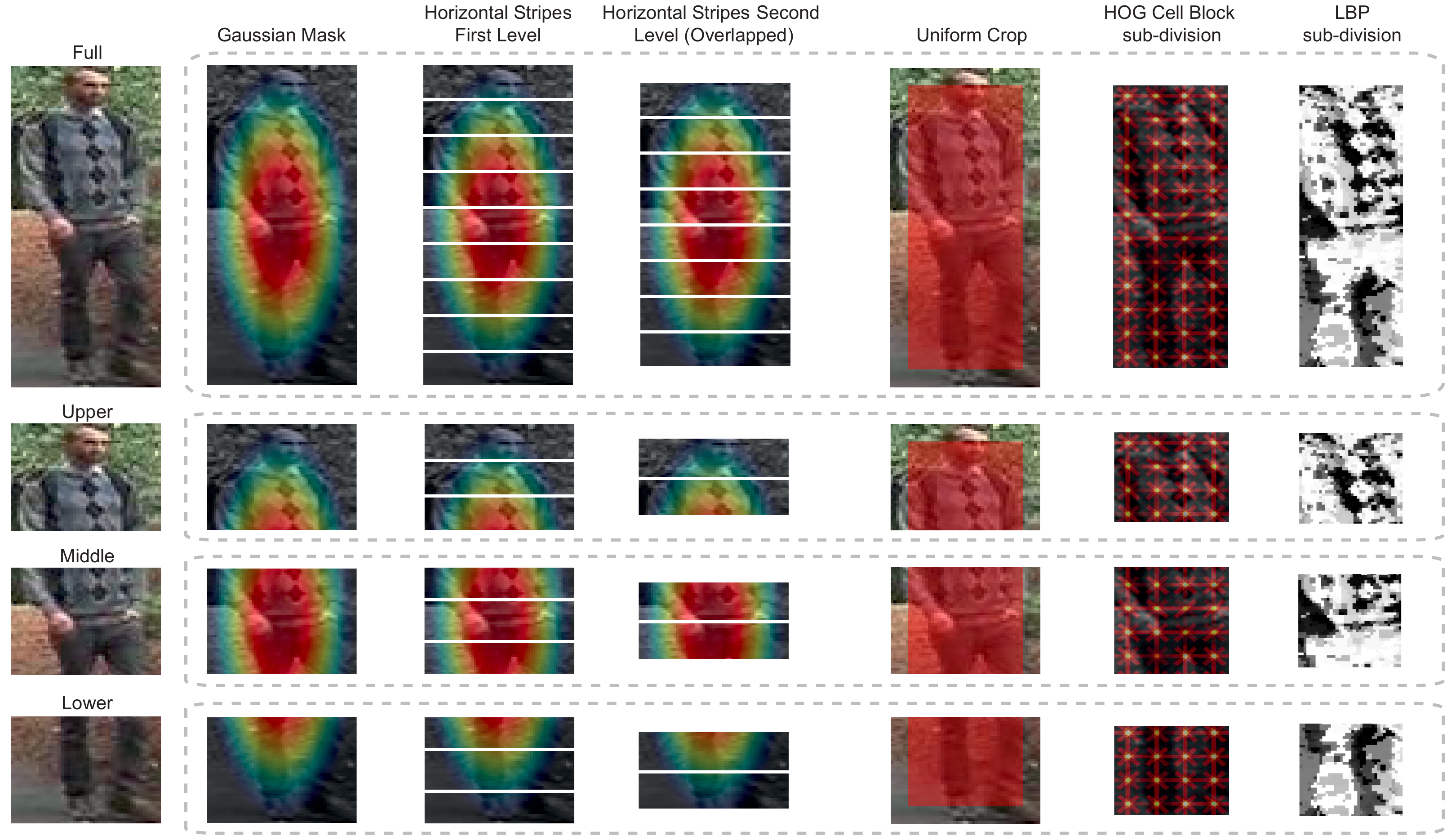}
   \caption{Illustration of our feature descriptor. We extract color
     (HS, RGB and Lab) and texture (HoG and LBP) features from the full image
     and from the upper, middle and lower regions of the image.}
\label{fig:whos2}
\end{figure*}

\section{Person Representation}
\label{sec:feature}

In order to account for spatial distribution of the person appearance, our representation model considers a coarse segmentation of the person image into upper, middle and lower image regions. 
Texture features are extracted for the whole image and each region and represented with HOG~\cite{dalal05} and Local Binary Pattern (LBP) histograms. For color, a finer segmentation into overlapping stripes is considered for each component (see Figure~\ref{fig:whos2}). Color information is modeled by histograms in the Hue Saturation, RGB and Lab color spaces, in order to account for differences in illumination due to different viewpoints.  

Separate channels are therefore maintained that model the person representation for each feature and each component separately, namely:
$$\{HS\}_{p},~\{RGB\}_{p },~\{Lab\}_{p},~\{HOG\}_{p},~\{LBP\}_{p} $$
where the suffix $p$ stands for the full ($f$), upper ($u$), middle ($m$) and lower ($l$) components of our representation.

The person images are resized to the resolution of $126\times64$ pixels. 
For color channels, the contribution of each pixel to each histogram bin is
weighted through a non-isotropic Gaussian kernel
to decrease background pixels influence without requiring an explicit background segmentation. 
For texture channels, we remove 6 pixels from the image border and compute HOG descriptor with 4-bin gradient orientation, and the 58-bin LBP.

\begin{figure*}[t]
\centering
\includegraphics[width=0.935\textwidth]{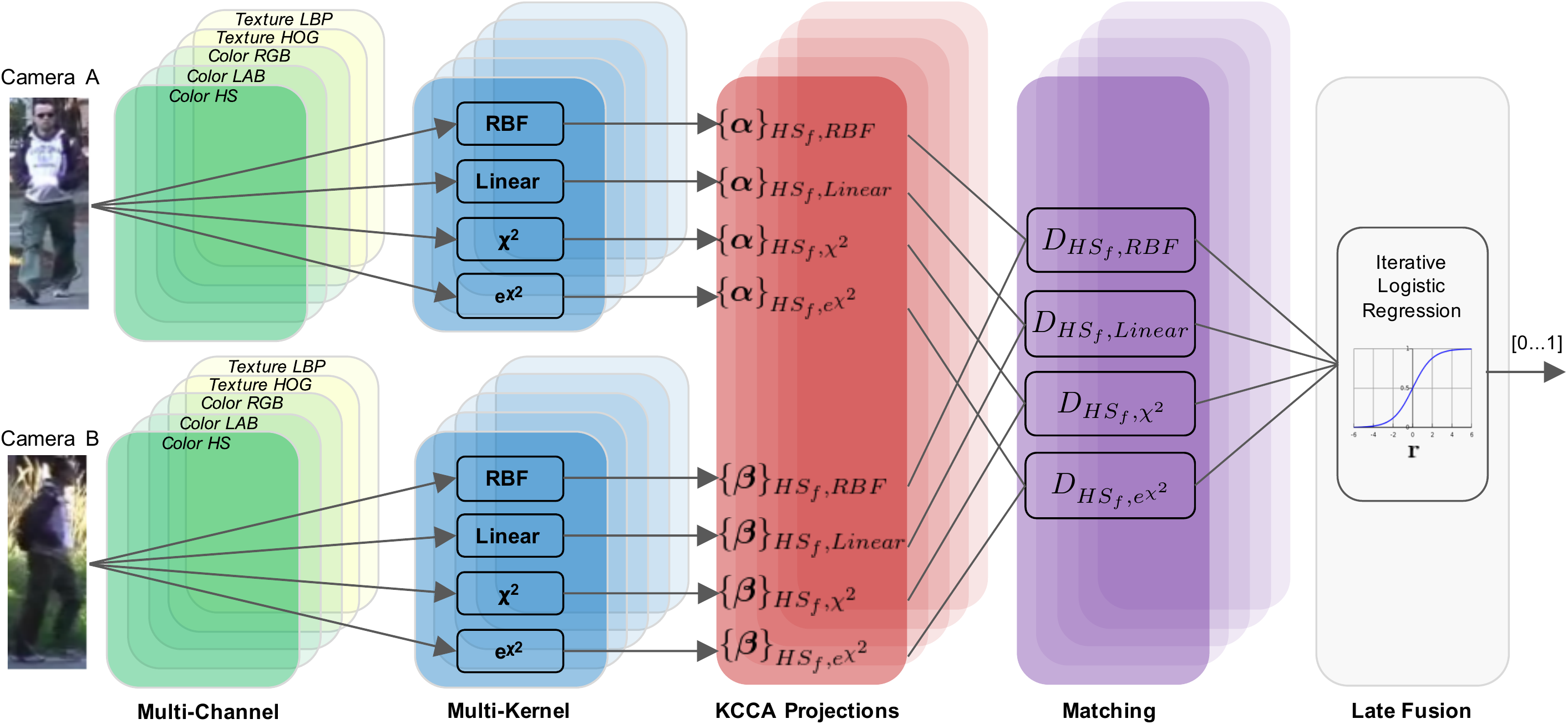}
   \caption{An illustrative figure of our multi-channel, multi-kernel KCCA approach.
     Each feature channel is fed to different kernels: for the sake of clarity we show a single channel $HS_f$ in the figure. For each of these 
     combinations, we learn specific KCCA projections and then use the
     learned projections to map each kernel-channel into its common
     subspace. Cosine distance is used to perform matching given a
     kernel-channel pair. Finally, distances coming from all the
     combinations are stacked together to form a final features vector.
     This feature vector is used to perform recognition through an 
     iterative logistic regression.}
\label{fig:illustrationMCKCCA}
\end{figure*}

\section{Multi Channel-Kernel Canonical Correlation Analysis}
\label{sec:MCKCCA}
We learn a common projection space from pairs of images of the same person taken from two different cameras. To this end we employ KCCA~\cite{hardoon-2004} applied to each channel of our representation. 

We introduce the following notation. Given a feature channel $c$, being $\mathbf{F}_{a}(c)$ the set of feature vectors $\mathbf{f}_a(c)$ and $\mathbf{F}_{b}(c)$ the set of feature vectors $\mathbf{f}_b(c)$,  respectively for camera $a$ and $b$, and using camera $a$ for gallery and camera $b$ for probe,  we define: 
\begin{eqnarray}
\mathbf{F}_a(c) = \left[~ \mathbf{F}^T_{a}(c)~ | ~ \mathbf{F}^G_{a}(c) ~ \right]&\\
\mathbf{F}_b(c) = \left[~ \mathbf{F}^T_{b}(c)~ | ~ \mathbf{F}^P_{b}(c) ~ \right]
  \label{eq:trainingset}
\end{eqnarray}
where $\mathbf{F}^T_{a}(c)$ and $\mathbf{F}^T_{b}(c)$ are the training sets for the two cameras, $\mathbf{F}^G_{a}(c)$ is the gallery set of camera $a$ and $\mathbf{F}^P_{b}(c)$ is the probe set for camera $b$.

In the following, for the clarity of exposition, we will omit in the notations the reference to the channel $c$.

\subsection{Training KCCA}
\label{sec:KCCA}

KCCA constructs the subspace that maximizes the correlation between pairs of variables. Feature mapping into a higher-dimensional space is performed by exploiting the kernel trick. 

In our case, given corresponding feature vectors from the two cameras, for each channel, we denote $\mathbf{K}_a^{TT}$ and $\mathbf{K}_b^{TT}$ the kernel matrices of pairs from the training sets, $\mathbf{K}_a^{GT}$ the kernel matrix of pairs from the gallery and training sets, and $\mathbf{K}_b^{PT}$ the kernel matrix of pairs from training and probe sets respectively.  

The objective of KCCA is then to identify the projection weights
$\boldsymbol{\alpha},\boldsymbol{\beta}$ by solving:
\begin{equation}
\arg\max_{\boldsymbol{\alpha},\boldsymbol{\beta}}
\frac{\boldsymbol{\alpha}' \mathbf{K}^{TT}_a \mathbf{K}^{TT}_b \boldsymbol{\beta}}{\sqrt{\boldsymbol{\alpha}'
    {\mathbf{K}^{TT}_a}^2 \boldsymbol{\alpha} \boldsymbol{\beta}'
    {\mathbf{K}^{TT}_b}^2 \boldsymbol{\beta}}}.
\end{equation}
The norms of the projection vectors $\boldsymbol{\alpha}$ and $\boldsymbol{\beta}$ are regularized in order to avoid trivial solutions according to~\cite{hardoon-2004}.

The top $M$ eigenvectors of the
standard eigenvalue problem obtained after regularization provide the
final gallery and probe data projections: 
\begin{eqnarray}
\mathbf{\tilde{F}}^G_{a} & = & \mathbf{K}^{GT} \boldsymbol{\alpha} \cdot \boldsymbol{\lambda} \\
\mathbf{\tilde{F}}^P_{b} & = &\mathbf{K}^{PT} \boldsymbol{\beta} \cdot \boldsymbol{\lambda} 
\end{eqnarray}
where
\begin{equation}
\boldsymbol{\alpha}=\left[\boldsymbol{\alpha}^{(1)}\ldots \boldsymbol{\alpha}^{(M)}\right] \nonumber, 
\boldsymbol{\beta}=\left[\boldsymbol{\beta}^{(1)}\ldots \boldsymbol{\beta}^{(M)}\right]
\end{equation}
are the learned projections and  $\boldsymbol{\lambda}$ is the vector of eigenvalues obtained from KCCA. This gives more relevance to those dimensions in the projected space that have higher eigenvalues, so improving the overall matching performance.

In order to improve re-identification, we compute KCCA with four different kernels, so to have multiple representations. Namely, we use a linear kernel, a Gaussian radial basis function kernel (RBF), a ${\chi}^2$ kernel and an exponential ${\chi}^2$ kernel. The distance matrix $D(\mathbf{\tilde{F}}_{a}^{G},\mathbf{\tilde{F}}_{b}^{P})$ can be computed, that defines for each kernel and feature channel the cosine distance between the gallery and probe images.

\section{Selection of the optimal kernel-channel}
\label{sec:weightedlatefusion}

According to our person representation, for each image pair, a distance vector of 80 values (four components with five features each, and four distinct kernels for KCCA) is defined.  Our goal is hence to select and weigh the most appropriate kernel and feature channels such that their combination results into the most effective re-identification. 
 
This is performed by taking subsets of $\mathbf{F}^T_a$ and $\mathbf{F}^T_b$ of equal size and calculating distances between KCCA projections through a two-fold cross validation. Learning of the relative distance weight for each combination  of kernel and feature channel is obtained with logistic regression. 
Distance weights $\mathbf{r}$ are learned through the optimization of the logistic regression function:
\begin{equation}
\underset{\mathbf{r}}{\min} \, \frac{1}{2} \mathbf{r}'\mathbf{r} +
C\sum_{i} \sum_{j} \log (1 + e^{- y_{ij} \mathbf{r}'
  \mathbf{d}^{TT}_{ij} } )
\label{eq:opt_lt}
\end{equation}
where $\mathbf{d}^{TT}_{ij}$ is the distance vector between the feature vector $\mathbf{\tilde{f}}^T_{a_i}$ (from camera $a$) and the feature vector ${\tilde{\mathbf{f}}}^T_{b_j}$ (from camera $b$) after KCCA projection with a \emph{bias} term; $y_{ij} = \{-1,1\}$ accounts for the fact that the two features correspond to the same person in the two views, and $C$ is a penalty parameter (in the experiments set equal to 1). 

Positive weights indicate a non reliable distance obtained from a combination of kernel and feature channel. In fact, a large kernel-feature distance, that should correspond to a non matching pair, combined with a positive weight can actually lead to a high matching probability.
According to this observation, we derive an iterative filtering procedure to progressively drop out kernel-feature channels that have positive weights and learn the logistic regression on the remaining distance vectors.

\subsection{Matching with iterative logistic regression}
Given the learned weights $\mathbf{r}$ from our iterative logistic regression, we 
have now a principled way to fuse all the distances $D(\mathbf{\tilde{F}}_{a}^{G},\mathbf{\tilde{F}}_{b}^{P})$ together at test time. 
The probability $\hat{p}(i,j)$ between the feature vector $\mathbf{\tilde{f}}^G_{a_i}$ (from camera $a$) and the feature vector ${\tilde{\mathbf{f}}}^P_{b_j}$ (from camera $b$) after KCCA projection to be the same person is calculated as:
\begin{equation}
\hat{p}(i,j) =  \frac{1}{1+ \exp(-\mathbf{r}'  \mathbf{d}^{GP}_{ij})}.
\label{eq:lr}
\end{equation}
The overall process is represented in Figure~\ref{fig:illustrationMCKCCA} and referred to as multi-channel, multi-kernel canonical correlation analysis (MCK-CCA) in the following experiments.

\section{Experiments}
\label{sec:experiments}

We run our experiments on four standard publicly available
datasets for re-identification that are VIPeR~\cite{elf:08}, 
PRID~\cite{RPLM:eccv12}, PRID 450s~\cite{roth14a} and CUHK01~\cite{Li2013}.

VIPeR~\cite{elf:08} presents illumination variations and pose changes between pairs of views. We split the whole
set of 632 image pairs randomly into two sets of 316 image
pairs, one for training and the other one for testing. The
testing set is further split into a gallery and a probe set. A single
image from the probe set is selected and matched with all the images
from the gallery set. The process is repeated for all images of the
probe set and the evaluation procedure is run on the 10 splits
publicly available from~\cite{bazzani:reidcvpr10}.

The recent PRID dataset~\cite{RPLM:eccv12} is generally considered being more
challenging than VIPeR. It includes distractors as well as strong 
illuminations changes across cameras.  Differently from VIPeR, in this dataset, the person images are acquired from above with similar poses. 
Camera view $a$ contains 385 persons, camera view $b$ contains 749 persons, 200 appearing in both views. These image pairs are randomly split into a training and a test set of equal size. For the evaluation, camera $a$ is used as probe and camera $b$ is used as gallery. Thus, each of the 100 persons in the probe set is searched in a gallery set of 649 persons (where 549 are distractors).

The PRID 450s~\cite{roth14a} has almost the same 
characteristics of PRID but does not include distractors. Therefore, despite of the differences in appearance, the experimental setting for this dataset is similar to the one of VIPER. This dataset contains 450 person image pairs captured by two cameras and image pairs are split in
225 for training and 225 for test.

The CUHK01~\cite{Li2013} dataset, also known as CUHK Campus dataset,
was captured with two camera views in a campus
environment. Differently from the previous datasets, images in this
dataset present high resolution. Persons are mostly captured in a frontal pose from the first camera, and in a profile pose in the second camera with low illumination variations. 
This dataset contains 971 persons, and each person has two images in each camera view. Camera $a$
captures the frontal view or back view of pedestrians, while camera $b$
captures the side views. The person identities are split into 485 for
training and 486 for test.  This datasets provides two evaluation
modality: single-shot, one sample per subject (SvsS) and also
multi-shot setting with two samples per subject (MvsM N=2).

The evaluations for VIPeR, PRID and PRID 450s are conducted following
a single-shot protocol. While on CUHK01 we perform both single- and
multi-shot experiments. All the experiments are averaged over 10 trials.

\begin{figure}[t]
\centering
\includegraphics[width=0.45\textwidth]{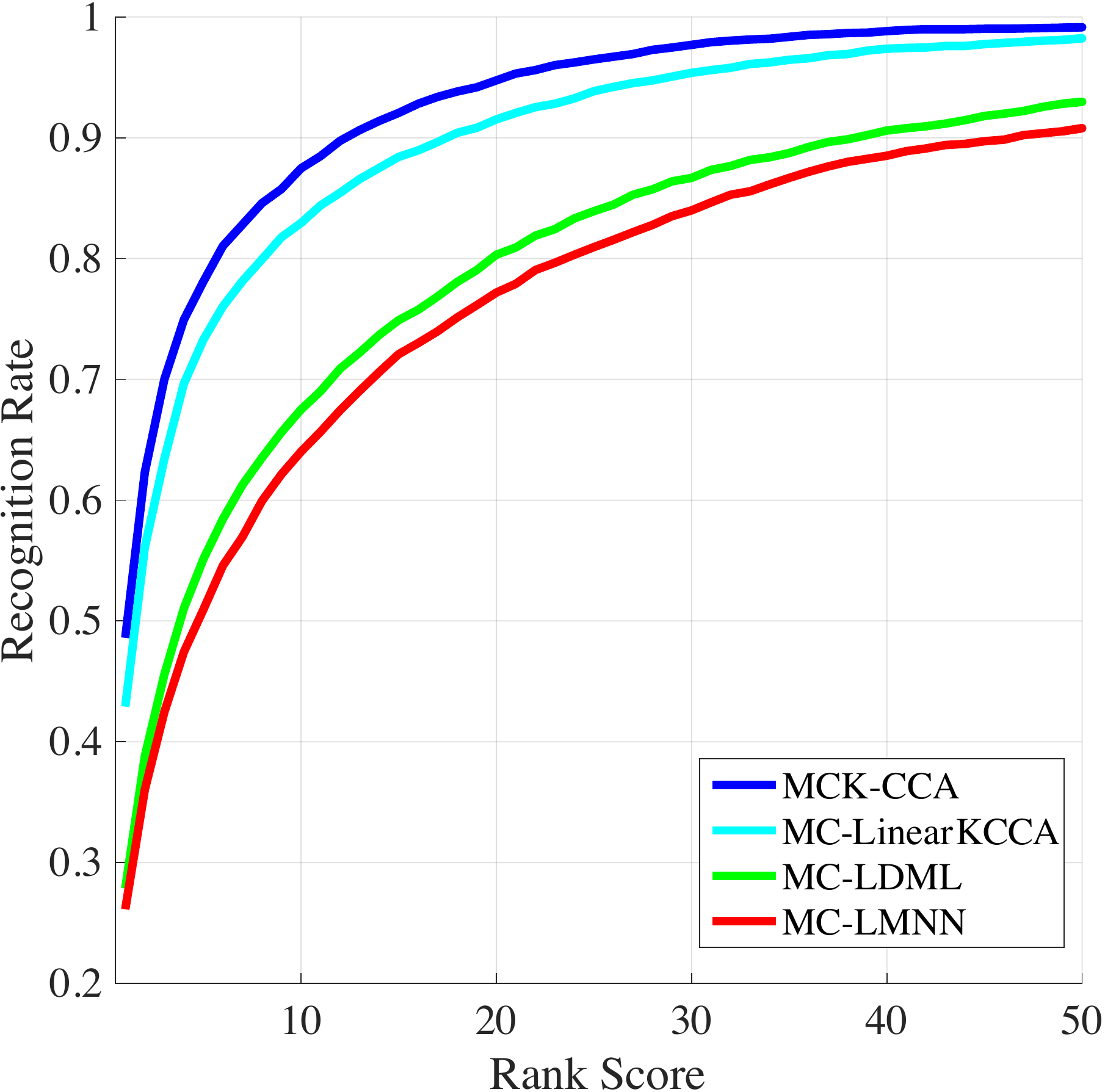}
   \caption{Comparison of MCK-CCA with metric learning methods LMNN and LDML on the VIPeR dataset.}
\label{fig:cmc_kissme}
\end{figure}

\begin{table}[t]
\centering
\resizebox{\columnwidth}{!}{
\begin{tabular}{|l||ccccc|}
\hline
\emph{Dataset} & \multicolumn{5}{c|}{VIPeR} \\
\hline
\emph{Rank} & 1  & 10 & 20  & 50  & 100 \\
\hline
\hline
EIML~\cite{EIML:avss12} & 22 & 63 & 78 & 93 & 98 \\
RPLM~\cite{RPLM:eccv12} & 27 & 69 & 83 & 95 & 99 \\
eSDC~\cite{eSDC:cvpr13} & 27 & 62 & 76 & -- & -- \\
KLMM~\cite{xiong2014person} & 28 &  76 & 88 & -- & -- \\
Li \emph{et al.}~\cite{Li_2013_CVPR} & 29.6 & 71 & 85 & 95 & -- \\
SalMatch~\cite{zhao2013:iccv13} & 30.1 & 65 & -- & -- & -- \\
PLS+prototype~\cite{PLS_prototype} &  33  & 78 &  87 & 96 & -- \\
Siamese CNN~\cite{Ahmed2015:CVPR} & 34.8 & 75 & -- & -- & -- \\
CSL~\cite{Shen_2015_ICCV} & 34.8 & 82.3  & 91.8 & 96.2 & -- \\
Improved DML~\cite{YiLL14:arxiv}  & 34.4 & 75.89 & 87.22 & 96.52 & -- \\
ECM~\cite{ECM_WACV15} & 38.9 & 78.4 & 88.9 & 96.0 & -- \\
SLTRL~\cite{SLTRL_Wang}  & 39.62 & 78.26 & 87.88 & -- & -- \\
LOMO+XQDA~\cite{Liao2015:CVPR} & 40.00 & 80.5 & 91.1 & -- & -- \\
Shi \emph{et al.}~\cite{Shi_2015_CVPR} & 41.6 & 86.2 & 95.1 & -- & -- \\
mFilter~\cite{Zhao2014:CVPR} & 29.11 & 65 & 80 & -- & -- \\
mFilter + LADF & 43.4  & 82 & 95 & -- & -- \\
Ensemble Metrics~\cite{Paisitkriangkrai_2015_CVPR} & 45.9 & \textbf{88.9} & \textbf{95.8} & \textbf{99.5} & \textbf{100} \\
KCCA $e^{\chi^2}$~\cite{lisanti:icdsc14} & 36.84  & 84.49  & 92.31 & 98.61 & 99.78\\
\hline
\hline
MCK-CCA with LR & 46.46 & 86.80 & 93.61 & 98.83 & 99.87\\
\hline
MCK-CCA with filteredLR & \textbf{47.85} & 87.34 & 93.83 &  98.89 & 99.78\\
\hline
\end{tabular}
}
\caption{Comparative performance analysis at ranks \{1,10,20,50,100\}
  with respect to the state-of-the-art on VIPeR.} 
\label{tab:sota_viper}
\end{table}

\begin{table*}
\centering
\begin{minipage}{0.48\textwidth}
\centering
\resizebox{\columnwidth}{!}{
\begin{tabular}{|l||ccccc|}
\hline
\emph{Dataset} & \multicolumn{5}{c|}{PRID (with distractors)} \\
\hline
\emph{Rank} & 1  & 10 & 20  & 50  & 100 \\
\hline
\hline
EIML~\cite{EIML:avss12} & 15 & 38 & 50 & 67 & 80 \\ 
RPLM~\cite{RPLM:eccv12} & 15 & {42} & {54} & 70 & 80 \\ 
Improved DML~\cite{YiLL14:arxiv} & 17.9 & 45.9 & 55.4 & 71.4 & -- \\ 
Ensemble Metrics~\cite{Paisitkriangkrai_2015_CVPR} & 17.9 & 50 & 62 & -- & -- \\ 
KCCA $e^{\chi^2}$~\cite{lisanti:icdsc14} & 16.55 & 49.25 & 61.00 & 78.55 & 89.85\\
\hline
\hline
MCK-CCA with LR & 26.10  &  61.10 &  \textbf{73.50} &  86.70 &  93.60 \\
\hline
MCK-CCA with filteredLR & \textbf{26.70} &  \textbf{62.10} &  73.30 &   \textbf{86.90} & \textbf{94.30} \\
\hline
\end{tabular}
}
\caption{Comparative performance analysis at ranks \{1,10,20,50,100\}
  with respect to the state-of-the-art on PRID. } 
\label{tab:sota_prid}
\end{minipage}
\quad
\begin{minipage}{0.48\textwidth}
\centering
\resizebox{\columnwidth}{!}{
\begin{tabular}{|l||ccccc|}
\hline
\emph{Dataset} & \multicolumn{5}{c|}{PRID 450s} \\
\hline
\emph{Rank} & 1  & 10 & 20  & 50  & 100 \\
\hline
\hline
SCNCD~\cite{Yang2014} &  26.9  & 64.2 &  74.9 & 87.3 & -- \\
PLS+prototype~\cite{PLS_prototype} &  28  & 63 &  75 & 89 & -- \\
ECM~\cite{ECM_WACV15} & 41.9 &  76.9 & 84.9  & 94.9 & --  \\
Shi \emph{et al.}~\cite{Shi_2015_CVPR} & 44.9 & 77.5 & 86.7 & -- & -- \\
CSL~\cite{Shen_2015_ICCV} & 44.4 & 82.2 & 89.8 &  96.0 & --\\
LOMO+XQDA~\cite{Liao2015:CVPR}  & 58.22 & 90.09 & \textbf{97.80}  & -- & -- \\
SLTRL~\cite{SLTRL_Wang}  & \textbf{59.38} & 88.68 & 94.67 & -- & -- \\
KCCA $e^{\chi^2}$~\cite{lisanti:icdsc14} & 38.09 & 81.33 & 90.44 & 97.91 & 99.73\\
\hline
\hline
MCK-CCA with LR & 55.42 & 90.36 & 95.16 & 98.49 & 99.78\\
\hline
MCK-CCA with filteredLR & 55.78 & \textbf{90.76} &  95.51 & \textbf{98.62} & \textbf{99.91} \\
\hline
\end{tabular}
}
\caption{Comparative performance analysis at ranks \{1,10,20,50,100\}
  with respect to the state-of-the-art on PRID 450s. } 
\label{tab:sota_prid450s}
\end{minipage}

\end{table*}

\begin{table*}
\centering
\begin{minipage}{0.48\textwidth}
\centering
\resizebox{\columnwidth}{!}{
\begin{tabular}{|l||ccccc|}
\hline
\emph{Dataset} & \multicolumn{5}{c|}{CUHK01 single-shot (SvsS)} \\
\hline
\emph{Rank} & 1  & 10 & 20  & 50  & 100 \\
\hline
\hline
DeepReId (FPNN)~\cite{Li_2014:CVPR}(*) & 27.8 & 73 & 89 & 95 & -- \\
Shi \emph{et al.}~\cite{Shi_2015_CVPR} & 31.5  & 65.8  & 77.6 & -- & --\\
Siamese CNN~\cite{Ahmed2015:CVPR} & 47.5 & 80 & -- & -- & -- \\
Ensemble Metrics~\cite{Paisitkriangkrai_2015_CVPR} & 51.9 & 83.0 & 89.4 & 95.9 & \textbf{98.6} \\
KCCA $e^{\chi^2}$~\cite{lisanti:icdsc14} & 38.11 & 74.22 & 82.37 & 92.04 & 95.78\\
\hline
\hline
MCK-CCA with LR & 55.76 & 86.38 & 91.81 & 96.52 & 98.48 \\
\hline
MCK-CCA with filteredLR & \textbf{56.61} & \textbf{86.79} & \textbf{91.98} & \textbf{96.65} & \textbf{98.58} \\
\hline
\end{tabular}
}
\caption{Comparative performance analysis at ranks \{1,10,20,50,100\}
  with respect to the state-of-the-art on CUHK01 single-shot. (*) Method not directly comparable because it uses 100 subject for test and 871 for training.} 
\label{tab:sota_cuhk01s}
\end{minipage}
\quad
\begin{minipage}{0.48\textwidth}
\centering
\resizebox{\columnwidth}{!}{
\begin{tabular}{|l||ccccc|}
\hline
\emph{Dataset} & \multicolumn{5}{c|}{CUHK01 multi-shot (MvsM N=2)} \\
\hline
\emph{Rank} & 1  & 10 & 20  & 50  & 100 \\
\hline
\hline
mFilter~\cite{Zhao2014:CVPR} & 34.3  & 65  & 74 & -- & --\\
SLTRL~\cite{SLTRL_Wang} & 61.6 & 90.2 & 94.4 & -- & 99 \\
LOMO+XQDA~\cite{Liao2015:CVPR} & 63.21 & 90 & 93 & -- & 99 \\
KCCA $e^{\chi^2}$~\cite{lisanti:icdsc14} & 47.70 & 84.26 & 90.82 & 96.32 & 98.40\\
\hline
\hline
MCK-CCA with LR & 66.42 & 92.04 & 96.24 & \textbf{98.70} & \textbf{99.67} \\
\hline
MCK-CCA with filteredLR & \textbf{69.49} & \textbf{93.07} & \textbf{96.19} & 98.56 & 99.65 \\
\hline
\end{tabular}
}
\caption{Comparative performance analysis at ranks \{1,10,20,50,100\}
  with respect to the state-of-the-art on CUHK01 multi-shot. } 
\label{tab:sota_cuhk01m}
\end{minipage}

\end{table*}

\subsection{Comparison with metric learning techniques}
In this experiment we compare our strategy to face the variety of person appearance in re-identification learning projection spaces where features from two cameras correlate using KCCA,  against metric learning applied to our feature model. 
In particular, we compare Large Margin Nearest Neighbor (LMNN)~\cite{weinberger2009distance}  and
Logistic Discriminant-based Metric Learning (LDML)~\cite{GVS09} techniques with our
multi-channel, multi-kernel KCCA (MCK-CCA). 
The experimental setting is the following: in
all the methods we employ our person descriptor composed of five feature
with four components; then, for each channel as defined in Sect.~\ref{sec:feature}, we compute the
LMNN, LDML and KCCA projections. We finally use the proposed iterative logistic regression fusion scheme to fuse all the distances together.
This experiment is conducted on the
VIPeR dataset and the performances are averaged over ten trials. 
We report the performance of our MCK-CCA obtained using only the linear kernel (MC-Linear KCCA) and all the kernels (MCK-CCA).
In Figure~\ref{fig:cmc_kissme} we can see
how MC-KCCA with the linear kernel only, improves over the two metric learning methods, under the same setting. The MCK-CCA with all the kernels achieves even higher performance. 
This experiment demonstrates that learning two projections, one for each camera, to map the data in a common space where features of the same person are highly correlated is more effective than learning a single metric~\cite{weinberger2009distance,GVS09}. Moreover, this observation is also supported by other recent methods which also propose the idea of learning both a metric and a common space to handle cross-view matching~\cite{Liao2015:CVPR}.

\subsection{Comparison with the state-of-the-art}

We now compare the performance of our approach with state-of-the-art
methods. In particular we provide a side-by-side comparison of the
proposed multi-channel, multi-kernel KCCA (MCK-CCA) with recent state-of-the-art
techniques such as: EIML~\cite{EIML:avss12},
RPLM~\cite{RPLM:eccv12}, eSDC~\cite{eSDC:cvpr13},
KLMM~\cite{xiong2014person}, SalMatch~\cite{zhao2013:iccv13},
PLS+prototype~\cite{PLS_prototype}, Siamese CNN~\cite{Ahmed2015:CVPR},
CSL~\cite{Shen_2015_ICCV}, Improved DML~\cite{YiLL14:arxiv},
ECM~\cite{ECM_WACV15}, SLTRL~\cite{SLTRL_Wang},
LOMO~\cite{Liao2015:CVPR}, Semantic-Attribute~\cite{Shi_2015_CVPR},
mFilter~\cite{Zhao2014:CVPR}, Ensemble
Metrics~\cite{Paisitkriangkrai_2015_CVPR}, SCNCD~\cite{Yang2014}.
For our method we both consider the case in which Logistic Regression with and without iterative filtering of the kernel-feature channels is used (``MCK-CCA with LR'' and ``MCK-CCA with filteredLR'', respectively).

In table~\ref{tab:sota_viper} we report the results on the VIPeR
dataset. 
It is worth noticing that the Ensemble Metrics and our method, learning multiple metrics and projections respectively, to cope with the variations of pose and illumination of this dataset, score the best results.
It appears that our MCK-CCA improves of few percentage points at rank-1 with respect to the Ensemble Metrics.
The LOMO+XQDA~\cite{Liao2015:CVPR} method exploits metric learning and performs feature projection into a common space between the two views, as in our solution, although with a different method, but has a much lower performance. 
Finally, among the CNN-based methods the Siamese CNN~\cite{Ahmed2015:CVPR} has the best performance but clearly does not achieve a state of the art result.

In table~\ref{tab:sota_prid} we show the recognition rate at various ranks on
the PRID dataset. 
Our method largely has the best performance.  The use of multiple color features and multiple common projection spaces for each of them, permits dealing with the strong illumination differences between the views. All the solutions using a single representation appear to be less robust.
Ensemble Metrics achieves 17.9\% recognition rate at rank-1, less than our method of about 10\%. It is worth to notice that our previous method~\cite{lisanti:icdsc14} using a single feature and KCCA projection with a single kernel has comparable performance with Ensemble Metrics.

In table~\ref{tab:sota_prid450s} we report the results on the PRID
450s dataset. On this dataset our MCK-CCA has similar performance trend and comparable
scores with SLTRL~\cite{SLTRL_Wang} and LOMO+XQDA~\cite{Liao2015:CVPR}. Both methods aim at learning transformation of the input to cope with appearance and pose variations. The SLTRL and LOMO+XQDA methods outperform our method at rank-1. 

On the CUHK01 dataset we performed comparisons for single-shot (SvsS) and multi-shot (MvsM N=2) modalities. Results are presented  in table~\ref{tab:sota_cuhk01s} and table~\ref{tab:sota_cuhk01m},  respectively.  For both protocols the MCK-CCA scores the best results. The capability of KCCA to correlate different representations into a common projection space is clearly evident in this case.

\begin{figure*}[tb]
\centering
\subfigure[VIPER]{
\includegraphics[width=.18\textwidth]{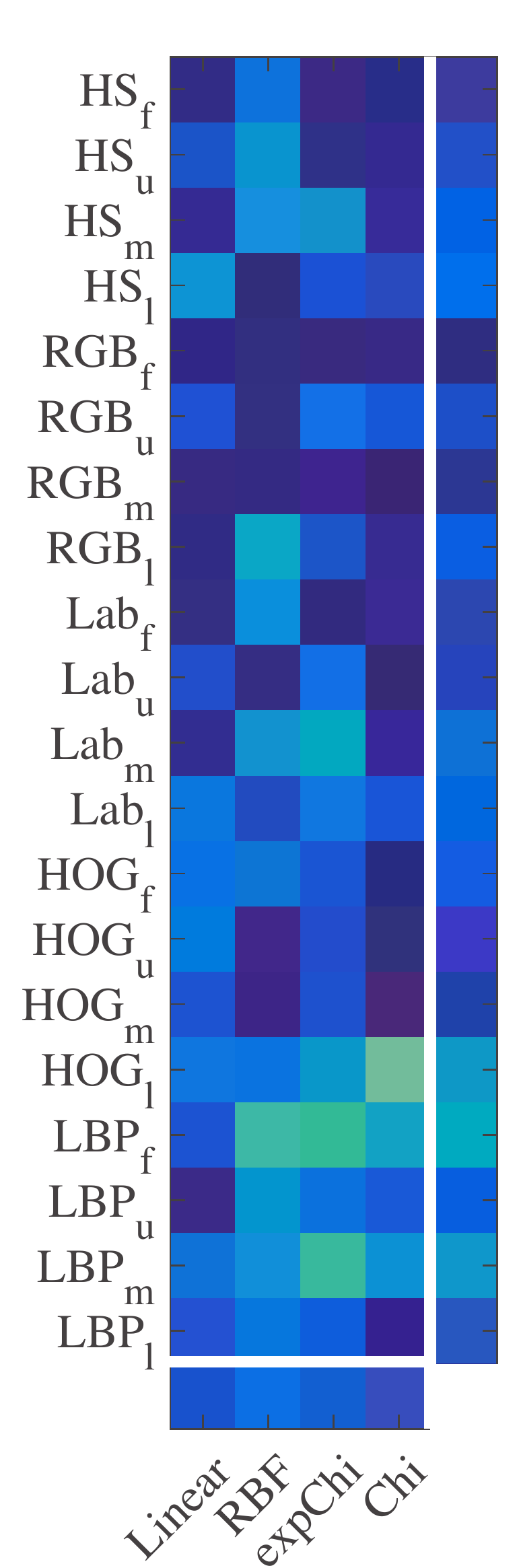}
}
\quad\quad\quad
\subfigure[PRID]{
\includegraphics[width=.134\textwidth]{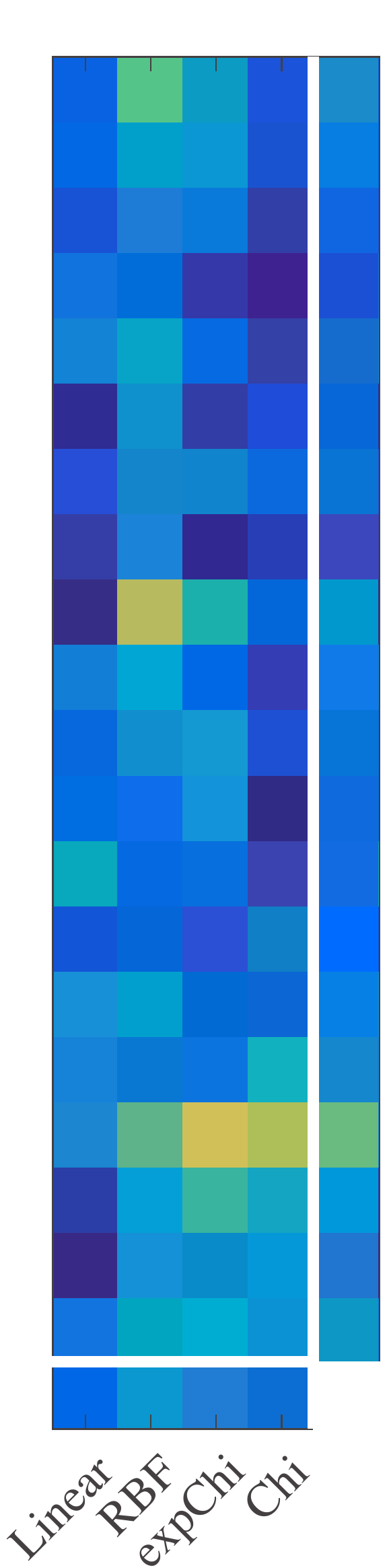}
}
\quad\quad\quad
\subfigure[PRID 450s]{
\includegraphics[width=.135\textwidth]{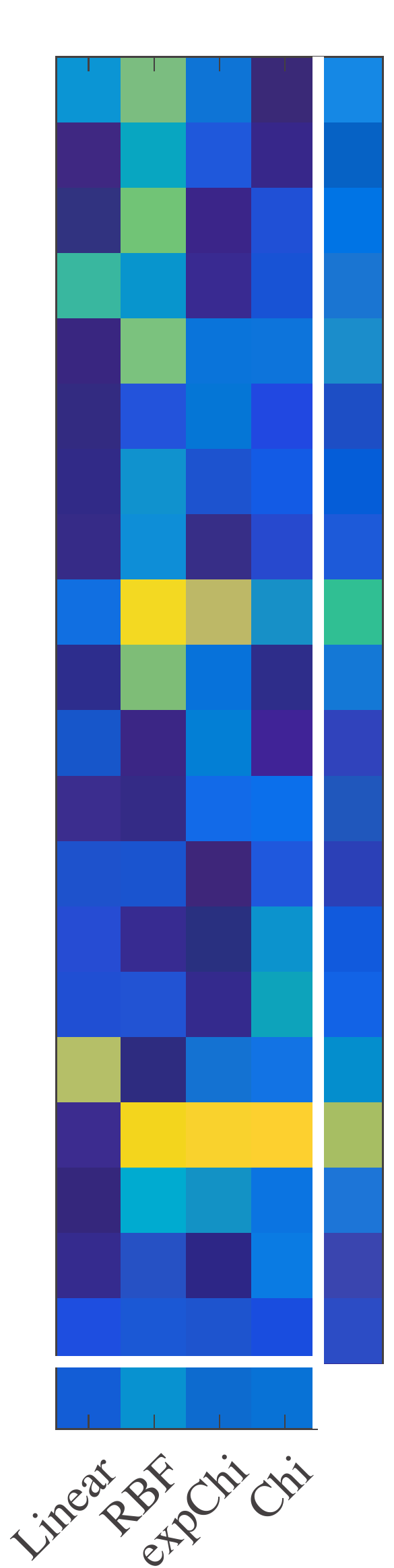}
}
\quad\quad\quad
\subfigure[CUHK01]{
\includegraphics[width=.135\textwidth]{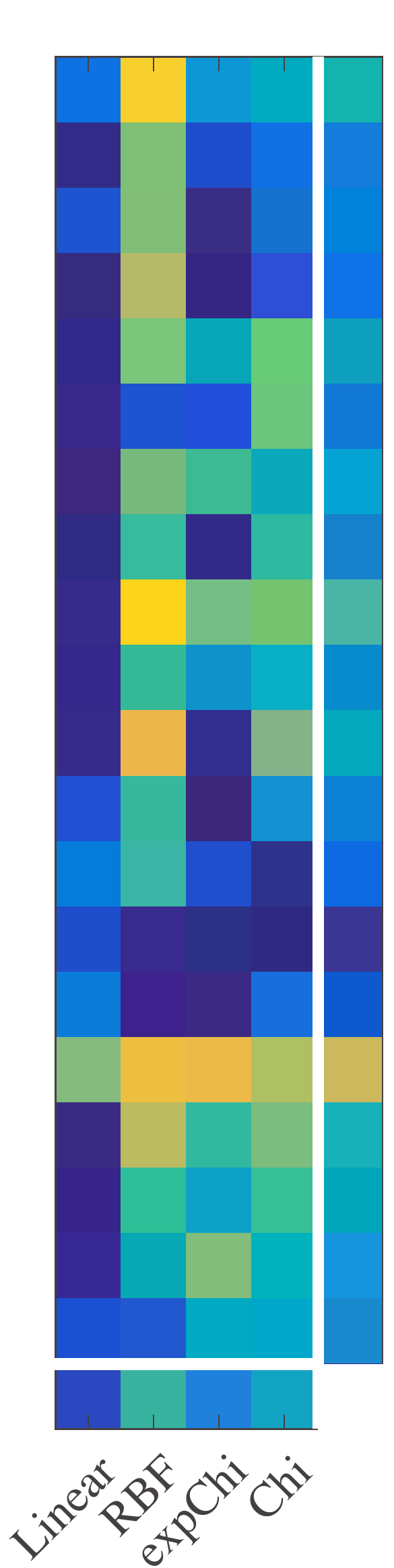}
}
\subfigure{
\includegraphics[width=.053\textwidth]{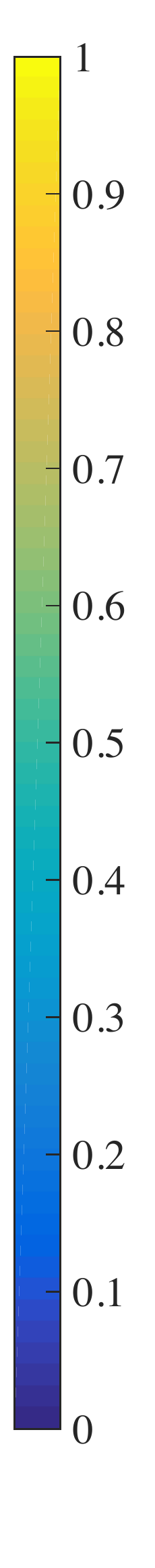}
}
\caption{Filtering analysis for each feature channel and kernel combination
  on the VIPeR, PRID 450s, PRID and CUHK01 datasets. Rows show channels, whereas columns show kernel.
 Each plot report also the summation over kernel (bottom) and the summation over channels (right).  
  }
\label{fig:filtering2}
\end{figure*}

\subsection{Analysis of iterative logistic regression}
\label{sec:logistic_analysis}

In this section, we analyze the performance of our iterative 
fusion scheme by giving insights on how each kernel and channel combination is
filtered out by our iterative logistic regression. 
Additionally, we show the
difference in performance between our MCK-CCA fusion scheme  
 with respect to an early fusion KCCA
baseline~\cite{lisanti:icdsc14}.

Each plot in Figure~\ref{fig:filtering2} shows the probability of weight
filtering per dataset: on the y-axes we report the channels, whereas on the
x-axes we report the kernels.
 Moreover, for each figure, on the right part of the matrix, we show how many times a given channel is
removed across all the kernels; instead, at the bottom, we show how
many times a kernel is removed, across all the channels. 
We can see that MCK-CCA makes an extensive use of the iterative
logistic regression filtering on all the datasets. VIPeR may be seen as an exception as most of the kernel-channels are 
often maintained.
Our analysis on VIPeR shows that weak channels are represented by HOG$_\text{l}$, LBP$_\text{f}$ and LBP$_\text{m}$. 
These channels correspond to texture features that may be noisy on VIPeR due to the low image resolution.

Regarding PRID, PRID 450s and CUHK01, we can
observe that in general for texture features, the less relevant components are
the full and lower region. Especially, the LBP$_\text{f}$ and HOG$_\text{l}$ are often filtered out. 
It is also possible to see that, despite being largely used in literature, the RBF kernel is dropped out more 
frequently than the other kernels, especially in the CUHK01 dataset. These three datasets
have in common that the channel Lab$_\text{f}$ is removed with high probability, while it is working
well on VIPeR.

Finally, considering all the results presented in 
tables~\ref{tab:sota_viper},~\ref{tab:sota_prid},~\ref{tab:sota_prid450s},~\ref{tab:sota_cuhk01s},~\ref{tab:sota_cuhk01m},
we can see that our late fusion outperforms, by a significant
margin, a single KCCA learned over the stacked features, as proposed in~\cite{lisanti:icdsc14}.  This is mainly
because of the fact that a late fusion schemes allows maximizing the
discriminative power of each kernel-channel combination. Moreover in
most of the cases the iterative logistic regression schemes is able to
select the most important kernel-channel combination and weigh them
in order to give more importance to the most discriminative ones.

\section{Conclusion}
\label{sec:conclusions}
We have presented a method to overcome one of the main
challenges of cross-view re-identification, that is dealing with drastic appearance changes. 
MCK-CCA grounds on the idea of addressing the extreme variability of person appearance in different camera views through multiple representations.
These representations are projected onto multiple spaces that emphasize appearance correlation using KCCA and different kernels.
Finally, our solution learns the most appropriate combinations for the observed pair through 
an iterative logistic regression, producing compelling results on standard re-identification benchmarks.
The proposed technique showed also to be competitive with state-of-the art method based on metric-learning.
Investigating the possibility of directly incorporating metric-learning technique into our approach could represent an interesting line of research for future works.

\section*{Acknowledgment}
\small
This research is partially supported by ``THE SOCIAL MUSEUM AND SMART TOURISM'', MIUR project
no. CTN01\_00034\_23154\_SMST.

\bibliographystyle{IEEEtran}

\begin{IEEEbiographynophoto}{Giuseppe Lisanti}
received the PhD degree in
computer science from the Universit\`a di Firenze.
He is a postdoc at the Media Integration and
Communication Center and his main research
interests focus on computer vision and pattern recognition, specifically
for person detection and tracking, person re-identification, 2D and 3D face recognition.
\end{IEEEbiographynophoto}

\begin{IEEEbiographynophoto}{Svebor Karaman}
received the PhD degree in
computer science from the University of Bordeaux, France. 
He is currently a postdoctoral researcher at Columbia University, NY, USA. 
His research interests include computer vision and machine learning, more specifically video analysis and indexing,
action recognition, person re-identification, face recognition and hashing.
\end{IEEEbiographynophoto}

\begin{IEEEbiographynophoto}{Iacopo Masi}
received the PhD degree in
computer science from the Universit\`a di Firenze, Italy. 
He is currently a postdoctoral scholar at University of
Southern California, USA. 
His research interests include pattern
recognition and computer vision, specifically the
subjects of tracking, person
re-identification, 2D/3D face recognition and modeling.
\end{IEEEbiographynophoto}

\end{document}